\documentclass[lettersize,journal]{IEEEtran}
\usepackage{amsmath,amsfonts}
\usepackage{algpseudocode} 
\usepackage{algorithm}
\usepackage{array}
\usepackage[caption=false,font=normalsize,labelfont=sf,textfont=sf]{subfig}
\usepackage{textcomp}
\usepackage{stfloats}
\usepackage{url}
\usepackage{verbatim}
\usepackage{graphicx}
\usepackage{cite}
\usepackage{xcolor} 

\begin{document}

\title{Soft Robotic Exogloves for Dexterous Mobility - Towards Personalized Rehabilitation}

\author{Paul Dela Cruz$^{1}$,~\IEEEmembership{Graduate Student Member,~IEEE,}, Mostafa Mo. Massoud$^{1}$,~\IEEEmembership{Graduate Student Member,~IEEE,} and Jacqueline Libby$^{2}$,~\IEEEmembership{Member,~IEEE}
\thanks{This work was supported by the US National Science Foundation under Grant 2502197. (\emph{Corresponding author: Jacqueline Libby})}%
\thanks{$^{1}$Paul Dela Cruz, Mostafa Mo. Massoud and Jacqueline Libby are with the Department of Mechanical Engineering, Stevens Institute of Technology, 1 Castle Point Terrace, Hoboken, NJ, 07030, USA
        {\tt\small (email: pdelacr1@stevens.edu; mmassoud@stevens.edu; jlibby@stevens.edu).}}%
}




\maketitle
\thispagestyle{empty} 

\begin{abstract}
Soft robotic exogloves can provide hand rehabilitation and assistance.
Fitting these gloves often relies on standardized measurements not tailored to the individual, limiting their effectiveness, especially for fine articulation necessary for dexterous manipulation.
We present the design, fabrication, modeling, and testing of a personalized pneumatically-actuated soft robotic exoglove.
The glove was fit to a user's hand with topological scans and fabricated with silicone mold casting.
Finite element analysis (FEA) was performed to evaluate actuator bending and forces from physical human-robot interaction (pHRI) between an actuator and a simplified personalized biomechanical finger model.
Pneumatic pressure control experiments were conducted to flex the user's finger with static and dynamic references.
Fabrication results show that topological scans enable precise tailoring to hand anatomy.
Simulations showed that anatomical personalization enables analysis of pHRI contact forces, and results indicate sufficient joint mobilization with non-ideal compression on the proximal phalanx.
Pneumatic testing indicates that pressure control allows accurate and targeted mobility of the metacarpophalangeal (MCP) and proximal interphalangeal (PIP) joints with intrinsic stiffness.
Testing of multiple designs showed that relaxing the strain-limiting layer improves actuator-to-finger joint alignment during actuation.
This work presents personalization to the human hand in structural conformability, joint topology, modeling of pHRI contact, and time-dependent actuation-deformation profiles.
This lays the groundwork for informing exoglove design optimization to enable assistance in dexterous manipulation and neuromuscular rehabilitation of fine motor skills.
\end{abstract}


\section{INTRODUCTION}
Rehabilitation robotics is an important tool for restoring motor function in individuals affected by neuromuscular diseases. 
Robotic systems can provide repeatable and task-oriented therapy, which can provide personalized therapy to improve functional recovery and motor learning~\cite{shahbazi2016robotics}. 
Soft robotic technologies further support these goals due to their inherent compliance and improved safety, which is more suitable for human-robot interaction \cite{cianchetti2018biomedical,Libby20230-ISMR}. 

Upper-limb rehabilitation presents challenges that differ from those of lower-limb systems. 
While lower-limb devices must address weight-bearing and dynamic stability, upper-limb rehabilitation must address precise joint coordination and fine motor control. 
Hand rehabilitation is particularly challenging, as there are many degrees of freedom in the hand and maintaining proper joint alignment during functional tasks can be difficult.
As a result, wearable robotic gloves are being introduced to assist finger motion in rehabilitation settings. 
Soft robotic gloves have been developed using a range of actuation strategies, including cable-driven and pneumatic systems. 
Cable-driven gloves mimic biological tendons to flex and extend the fingers \cite{park2016feasibility,park2018multimodal,palacios2023towards}. 
They can be lightweight and assist in grasping tasks suitable for activities of daily living \cite{xiloyannis2016modelling,alicea2021soft}. 
However, cable-driven systems often require complex routing and introduce friction and backlash, limiting scalability and ease of personalization. 
Pneumatically actuated soft gloves offer a simpler and more compliant alternative, making them well suited for wearable rehabilitation devices. 

Pneumatic soft exogloves have been implemented using fiber-reinforced \cite{tang2021model, heung2019design}, textile-based \cite{cappello2018assisting, zhou2019soft}, and elastomeric actuators \cite{polygerinos2013towards,tang2021probabilistic,Weigand2023-AMPA,heung2023assistive,Ridremont2024-Actuators}. 
Fiber-reinforced actuators have an elastomeric body and rely on fiber wrapping paired with a torque-compensating layer to enable bending \cite{polygerinos2015modeling,tang2020probabilistic}.
These actuators are often bulky and limited in their range of motion.
Textile-based actuators have a pneumatic bladder encapsulated in textiles where the elasticity of the fabric is controlled by the weave or knit of the fabric \cite{cappello2018assisting,sanchez2021textile}. 
They offer a lightweight option for soft robotic gloves but have complex fabrication processes that can hinder scalability \cite{sanchez20233d}. 
Elastomeric actuators such as the PneuNet actuators encode the bending behavior through geometric chamber design \cite{Shepherd2013-AdvMater}. 
The fast PneuNet (fPN) actuator design allows rapid bending and a greater range of motion, more suited for the fingers~\cite{Mosadegh2014-AdvFunctMater}.
Soft exogloves using these elastomeric actuators can target bending toward specific joints of the finger. 
In \cite{Ridremont2024-Actuators}, a soft exoskeleton was designed capable of bending joints of the wrist and fingers. 
Although these approaches have demonstrated effective assistance, these exogloves use standardized actuator geometries not tailored to individual hand anatomy.
This can lead to misalignment between actuator bending locations and human finger joints, potentially reducing comfort and effectiveness. 
There has been research in soft pneumatic exogloves with actuator geometries personalized to the hand; however, biomechanical measurements were obtained manually~\cite{tang2021probabilistic} or using 2D rectified images~\cite{Weigand2023-AMPA}. 

This work presents a pneumatically actuated soft robotic exoglove that is personalized to the anatomical dimensions of an individual’s hand using 3D scans. 
Actuator geometry and joint placement are designed based on hand measurements to align bending regions of the actuator with the metacarpophalangeal (MCP) and proximal interphalangeal (PIP) joints. 
The actuators have embedded strain-limiting layers and flex sensors to enable joint angle measurements.
A custom rigid dorsal base to secure the actuators was derived from the 3D hand scan to match the contours of the hand for consistent actuator placement.  

Additionally, this work presents a finite element analysis (FEA) to study the physical human-robot interaction (pHRI) between an individual's finger and the personalized actuator.
In \cite{Singh2023-AIM}, 3D scans are used to create a general biomechanical model of the wrist to study a soft actuator for wrist rehabilitation.
The work presented here uses 3D scans of an individual’s hand to create a biomechanical model of the finger, maintaining the proper anatomical dimensions.
Furthermore, we analyze the pHRI of two active degrees of freedom in series (MCP and PIP joints) for a better understanding of redundant multi-degree-of-freedom interactions between actuators and the human body.
This work allows us to consider the viability of fPN actuator joints as a functional module for soft robotic gloves. 

The remainder of this paper is organized as follows. 
Section~\ref{section.methods} presents the design and fabrication of the exoglove. Section~\ref{section.results} describes the FEA analysis and pneumatic experiments. Section~\ref{section.conclusions} discusses the findings and limitations that inform future work.

\section{Methods}\label{section.methods}

\subsection{Design}\label{subsection.design}

\begin{figure}[b!]
\centering
\includegraphics[width=.9\linewidth]{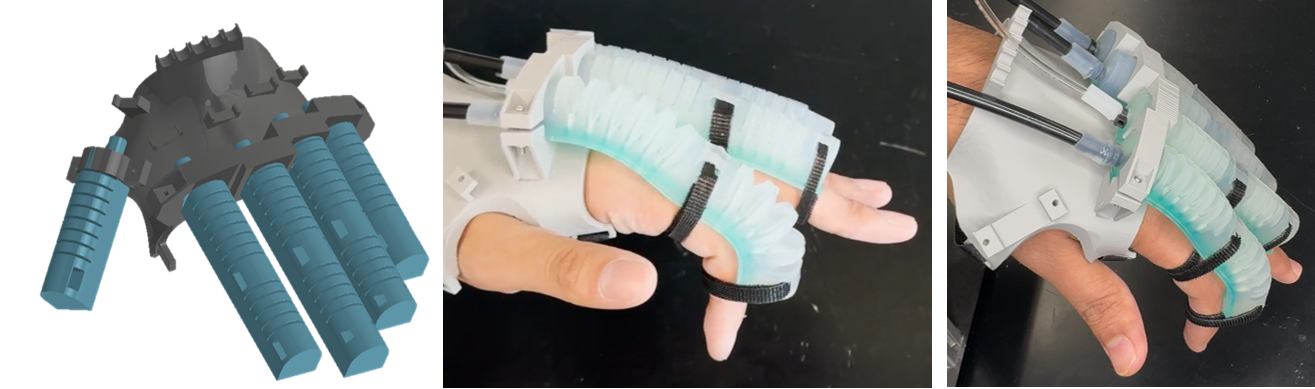}
\caption{Soft robotic exoglove presented in this work. \emph{(left)} CAD model of the glove. \emph{(Middle, right)} Real-world experiments with the fabricated glove, with the index finger robotically actuated.}
\label{fig.exoglove_intro}
\end{figure}

\begin{figure}[h!]
\centering
\includegraphics[width=.95\linewidth]{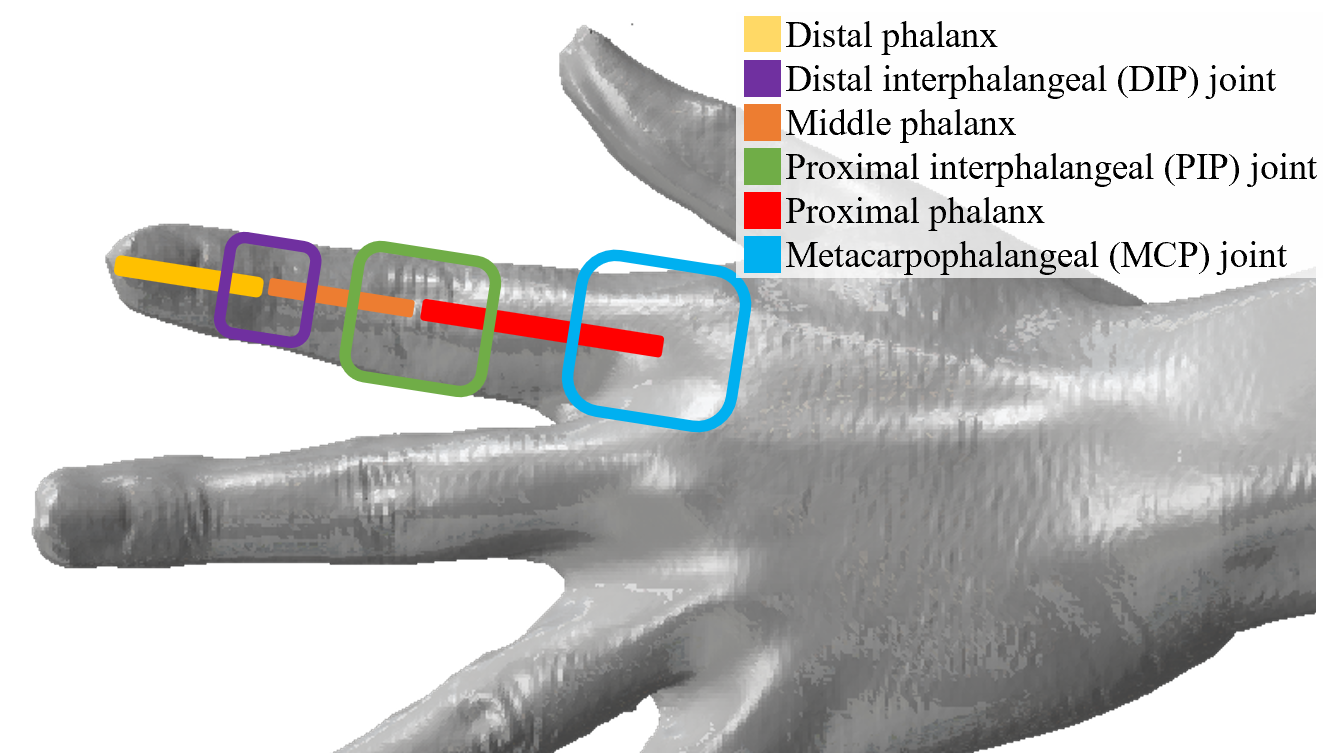}
\caption{The raw 3D scan of Participant A's left hand. The anatomy of the index finger is depicted with colors and described in the legend. Measurements from anatomical landmarks are used to customize each actuator's dimensions to each of the five fingers. The full contours of the dorsal side of the palm are used for customizing the base.}
\label{fig.rawScanDiagram}
\end{figure}

The soft robotic exoglove presented in this work is a personalized rehabilitation device to aid people suffering from neuromuscular diseases of the hand, shown in Fig.~\ref{fig.exoglove_intro}.
It consists of soft bending pneumatic actuators that control the movement of the fingers and a rigid base on the dorsal palm that secures the actuators.
An adjacent pneumatic control station is connected to the user with air tubing fed in through the base to pneumatically inflate each actuator.
The control station also feeds in electrical wiring to read measurements from flex sensors embedded in each actuator.
The actuators and base are fixed to the fingers and hand with straps.
The glove shown in Fig.~\ref{fig.exoglove_intro} was customized to Participant A, for whom informed consent was obtained.
A 3D scan of Participant A's hand, shown in Fig.~\ref{fig.rawScanDiagram}, was taken using the Artec Eva 3D scanner.
Measurements of each finger were taken to personalize the size of the actuators.
The contours of the hand scan were used to model the dorsal base.

\begin{figure}[thpb]
\centering
\includegraphics[width=1\linewidth]{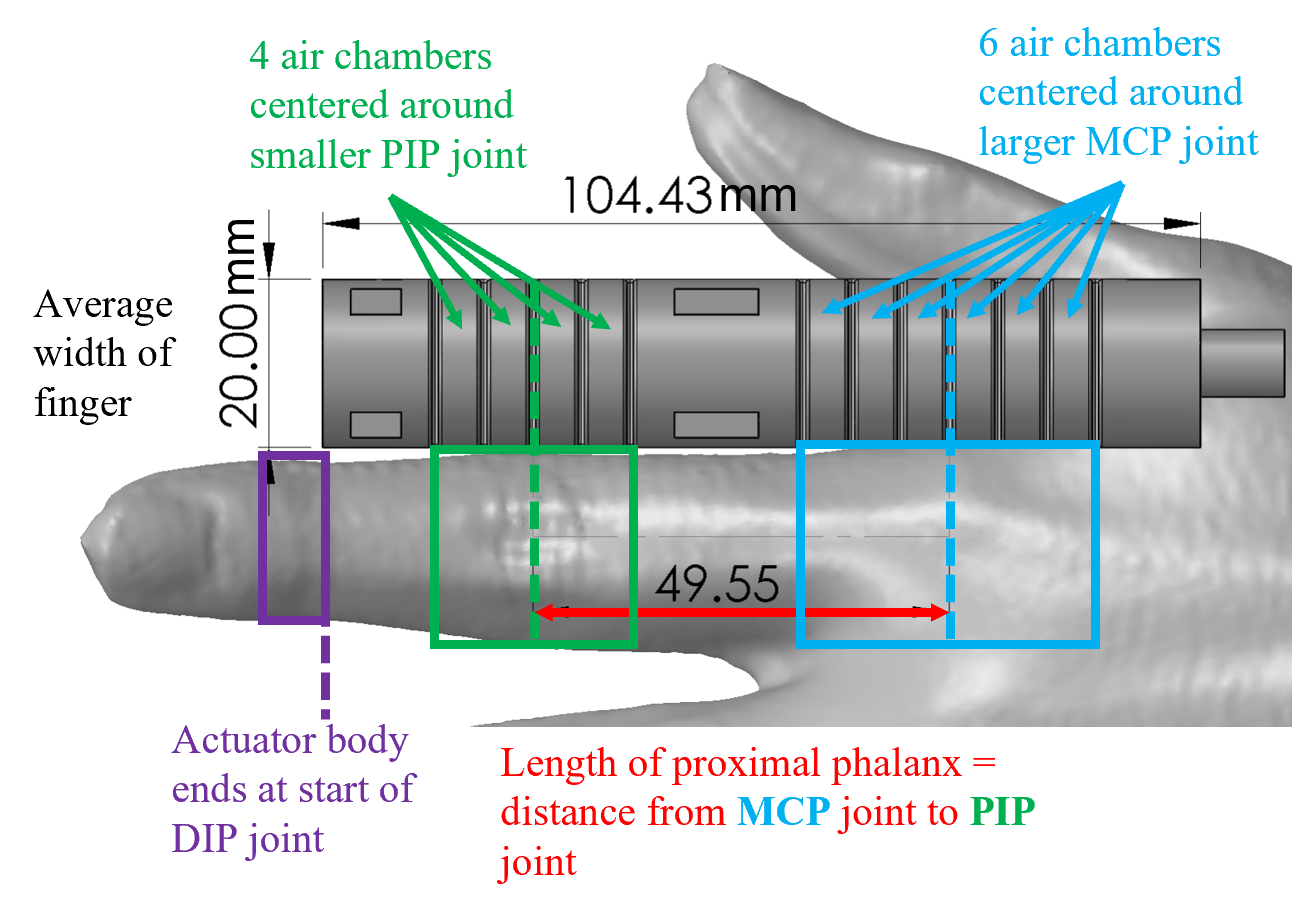}
\caption{The actuator features two joints: one to bend the PIP joint (green) and the other to bend the MCP joint (blue). The number of chambers of each joint of the actuator was based on the size of the corresponding joint of the finger. To personalize the size of the actuator to the hand, measurements of the proximal phalanx were taken to determine the distance between the two joints of the actuator.}
\label{fig.fingerMeasurement}
\end{figure}

\begin{figure}[thpb]
\centering
\includegraphics[width=.9\linewidth]{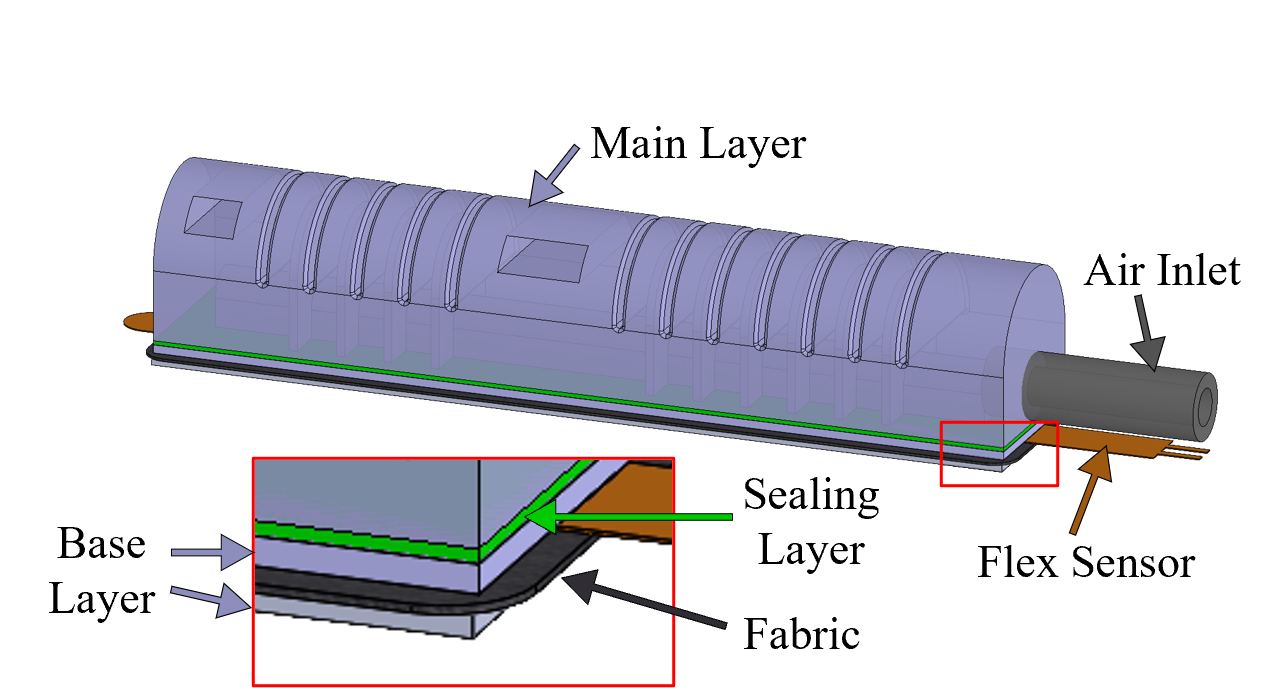}
\caption{The CAD model of the fast PneuNet (fPN) actuator. The actuator consists of three layers: main, base, and sealing. The main layer is the main body of the actuator, which contains the air chambers and air inlet. The base layer is where the fabric and flex sensor are embedded. The sealing layer is a thin layer of silicone to combine the main and base layers. The external gaps allow the PneuNet to bend and function properly.}
\label{fig.actuatorDiagram}
\end{figure}

\begin{figure}[thpb]
\centering
\includegraphics[width=.9\linewidth]{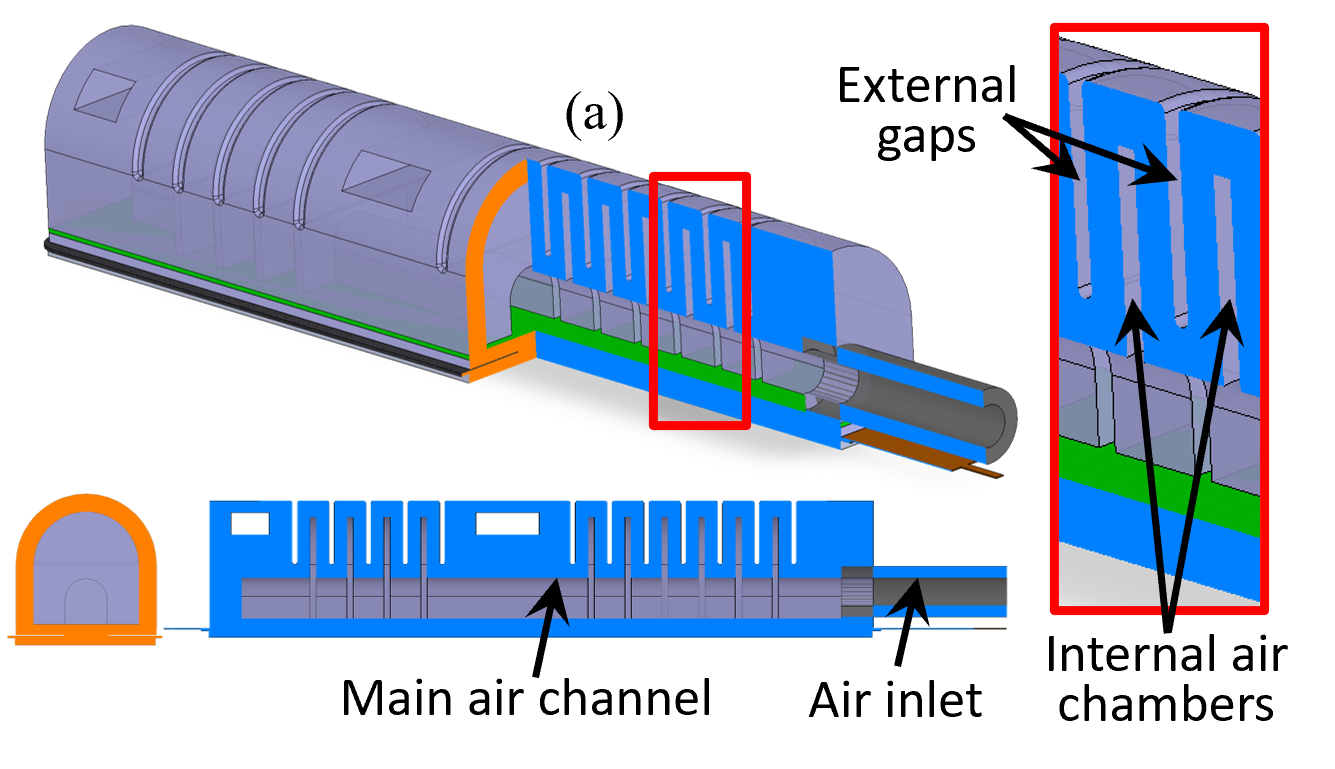}
\caption{The orange and blue show the sectional views of the actuator. The interior of the actuator has negative internal geometries, including the air inlet, the internal air chambers which expand to actuate the PneuNet, and the main air channel, which feeds air to the chambers. The orange and blue denote the solid silicone walls, to be distinguished from the negative space.}
\label{fig.actuatorSection}
\end{figure}

The exoglove soft bending actuators are based on the fast PneuNet design~\cite{Mosadegh2014-AdvFunctMater} where air chambers are separated by external gaps, enabling higher bending angles with less silicone deformation compared to other actuator designs.
This, in turn, allows for faster actuation.
Higher bending angles and faster actuation are steps towards enabling dexterous manipulation for fine motor control.
The actuators have two joints to focus bending at the MCP and PIP joints of the finger. 
The number of air chambers on each actuator joint is based on the size of the corresponding finger joint. 
As shown in Fig.~\ref{fig.fingerMeasurement}, the actuator joint responsible for bending the finger at the MCP joint has a series of six air chambers, whereas the smaller PIP joint has four air chambers. 
The length of the proximal phalanx determines the location of the MCP and PIP actuator joints to be centered with their corresponding finger joints.
The actuator body does not extend to the DIP joint.
For ease of fabrication, the actuator was split into three layers: the main layer, the base layer, and the sealing layer, as shown in Fig.~\ref{fig.actuatorDiagram}.
To measure the bending angle, a flex sensor is embedded in the base layer of each actuator. A cloth fabric is also embedded in the base layer as a strain-limiting layer to limit expansion in the axial direction.
The internal geometry of the actuators is shown in Fig.~\ref{fig.actuatorSection}. 

\begin{figure}[thpb]
\centering
\includegraphics[width=.8\linewidth]{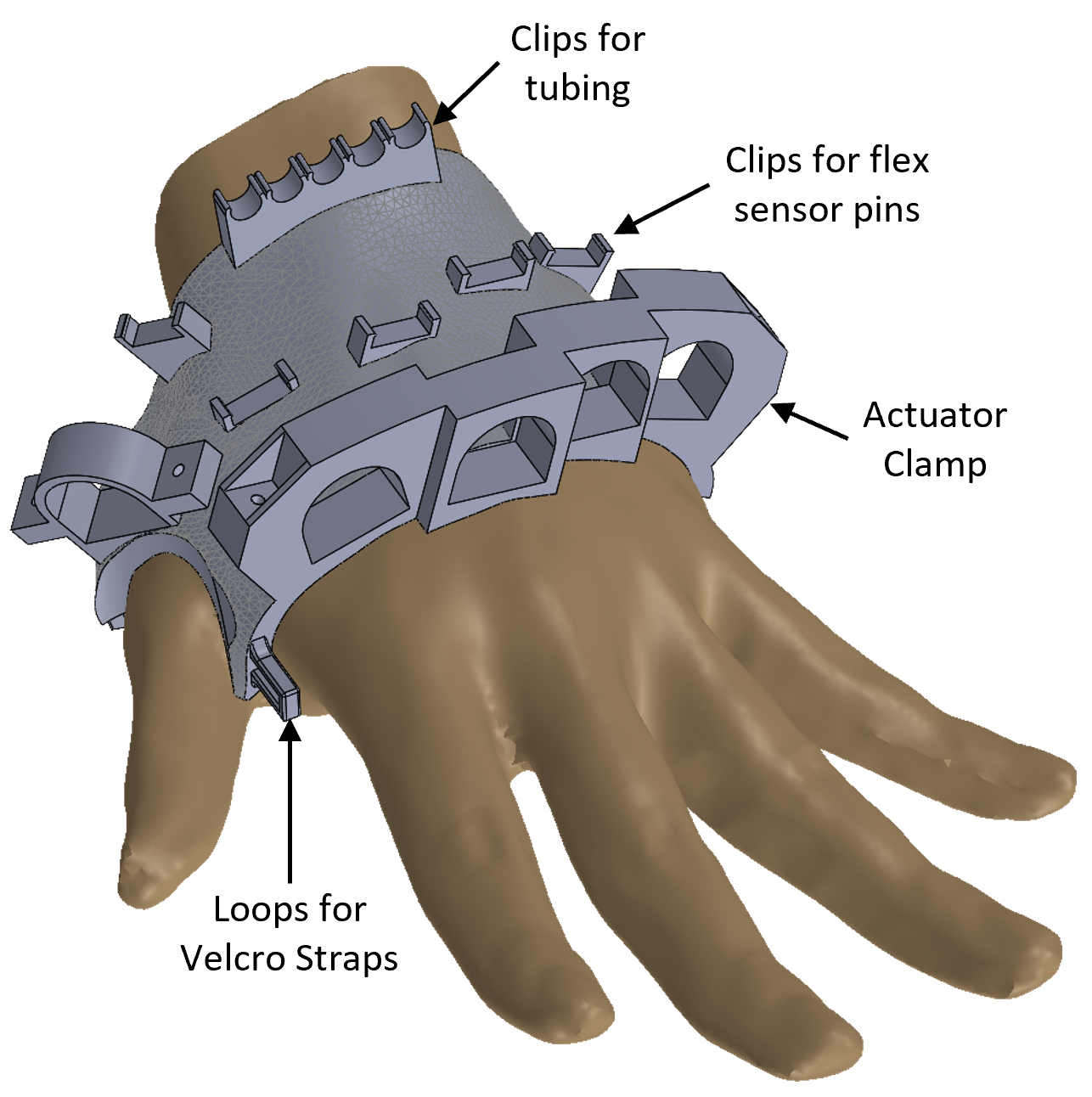}
\caption{The CAD model of the exoglove base, designed from the 3D scanned topology of the dorsal side of the hand.}
\label{fig.exoglove_base_CAD}
\end{figure}

The exoglove base, shown in Fig.~\ref{fig.exoglove_base_CAD}, was designed for customizability and electromechanical systems integration. 
It conforms to the contours of the dorsal hand based on the 3D scan.
The raw point cloud from the Artec Eva scanner was solidified and cropped in Meshmixer.
It was then imported into SolidWorks as an STL file. 
The hand model, excluding the fingers, was enclosed by a solid block and subtracted away, creating a negative imprint of the hand contours.
The resulting block was then duplicated and uniformly scaled up to define the base thickness.
The original and scaled blocks are aligned by their geometric centers.
Then a Boolean intersection is performed, retaining only the overlapping volume between the blocks.
This produced a shell that accurately captures the contours of the hand.
It was then trimmed down to the desired shape to fit the dorsal hand.

The exoglove base design also has features for integrating the electrical and pneumatic components with the human hand.
Clamps are added to the base such that the actuators are anchored and aligned to the fingers.
Clips are placed to secure and organize the tubing and wiring.
Loops for Velcro allow straps to secure the palm of the hand to the base.

\subsection{Fabrication}

\begin{figure}[thpb]
\centering
\includegraphics[width=.85\linewidth]{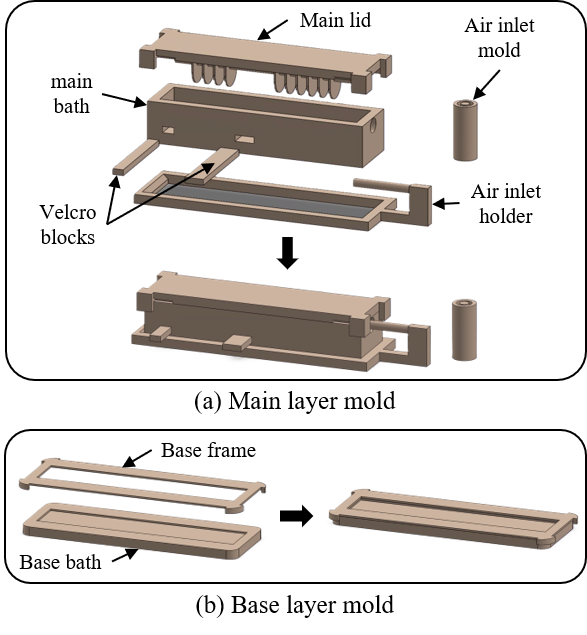}
\caption{Exploded CAD assemblies of the molds used to fabricate the actuators. (a) Main layer mold. (b) Base layer mold.}
\label{fig.moldDiagram}
\end{figure}

The exoglove base was 3D printed out of standard polylactic acid (PLA) plastic using a Bambu Lab X1 Carbon FDM printer with 15\% infill.
Molds were used to fabricate the actuators.
A diagram of the molds is shown in Fig.~\ref{fig.moldDiagram}.
The positive CAD model of the actuator presented in Section~\ref{subsection.design} was used to create the molds for the main and base layers.
The sealing layer is a thin layer of silicone used to combine the main layer with the base layer, and does not require a separate mold.
The molds were 3D printed out of PLA with 15\% infill using the Bambu Lab X1.
The silicone used to fabricate the actuators was Sorta Clear 40 (Smooth-On, Inc.), a two-part platinum-cure silicone rubber with a Shore hardness of 40A and a tensile strength of 800 psi. 
The 40A hardness, which is relatively high compared to common hyperelastic silicone material choices, was chosen to withstand a higher internal pressure and exert more force to bend the finger.
Silc Pig (Smooth-On, Inc.) is used to color the sealing layer green for troubleshooting.

\begin{figure}[thpb]
\centering
\includegraphics[width=.9\linewidth]{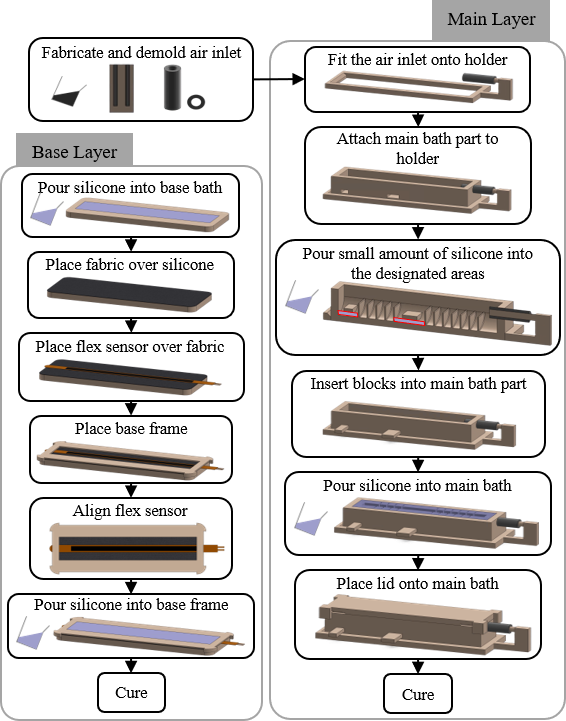}
\caption{A flow chart of the first half of the actuator fabrication process. Steps are shown for creating the air inlet, main layer, and base layer of the actuator.}
\label{fig.fabricationFC_A}
\end{figure}

The first half of the fabrication process to cast the main and base layers is shown in Fig.~\ref{fig.fabricationFC_A}.
The air inlets are fabricated and cured first, as they will be affixed to the main layer mold before pouring the main layer.
The air inlet is fitted onto the air inlet holder, which secures it in the proper orientation.
The main bath part is then attached to the holder.
Silicone is poured into the bottom mold to the height of the Velcro blocks before they are inserted to prevent air entrapment.
After inserting the blocks, the remaining silicone is poured into the mold, and the lid is placed on top.
For the base layer, silicone is first poured into the base bath.
Once the silicone settles, the cotton fabric, cut to be slightly larger than the base, is placed on top.
Tension is applied to the sides of the fabric so that it lies flat.
After the fabric becomes saturated with silicone, the flex sensor is placed.
Silicone is spread across the bottom of the flex sensor before it is placed in the middle of the fabric.
The base frame is dimensioned with a tolerance such that it snaps onto the base bath and excess fabric to clamp into place.
The flex sensor is then adjusted lengthwise for proper alignment with the pin mounts on the glove base.
Then silicone is poured into the base frame to finish the base layer.
The steps for the main and base layers can be performed and cured in parallel. 

\begin{figure}[thpb]
\centering
\includegraphics[width=.7\linewidth]{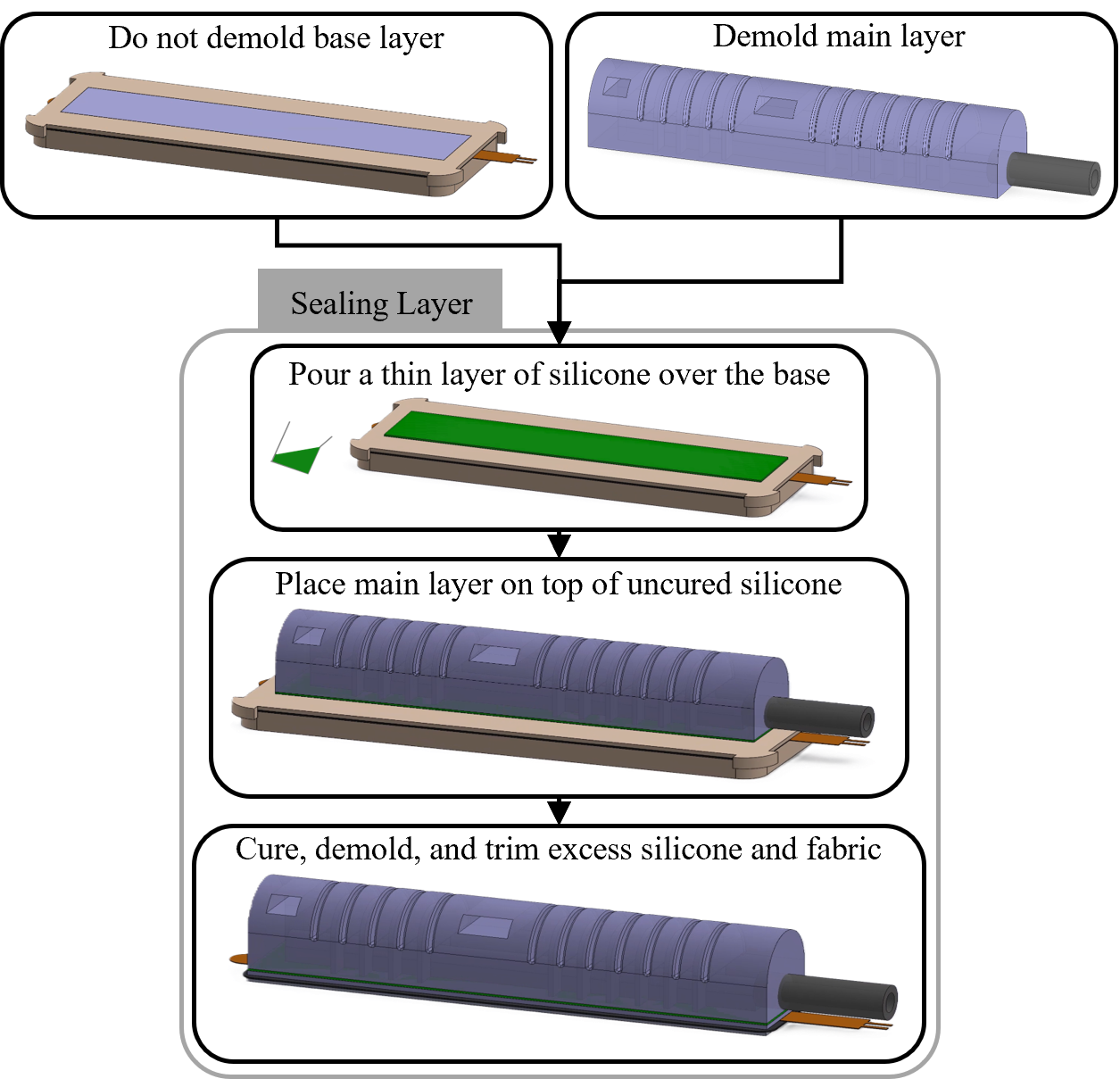}
\caption{A flow chart of the second half of the actuator fabrication process. Steps are shown for combining the main and base layers with the sealing layer.}
\label{fig.fabricationFC_B}
\end{figure}

The steps to combine the main and base layers are shown in Fig.~\ref{fig.fabricationFC_B}. 
After curing, the main layer is demolded, and excess silicone is trimmed. 
The base layer is left in its mold, allowing the sealing layer to be poured on top, and the main layer can be aligned with the base frame.
For the sealing layer, a small amount of silicone is poured and flattened with a craft stick. 
When the silicone has settled, the main layer is placed on top of the uncured silicone.
Once the sealing layer is cured, the actuator is demolded.
Pneumatic tubing is then inserted into the air inlet and secured with a miniature nylon cable-tie.

\section{Analysis and Results}\label{section.results}

\subsection{Calibration}

\begin{figure}[thpb]
\centering
\includegraphics[width=.9\linewidth]{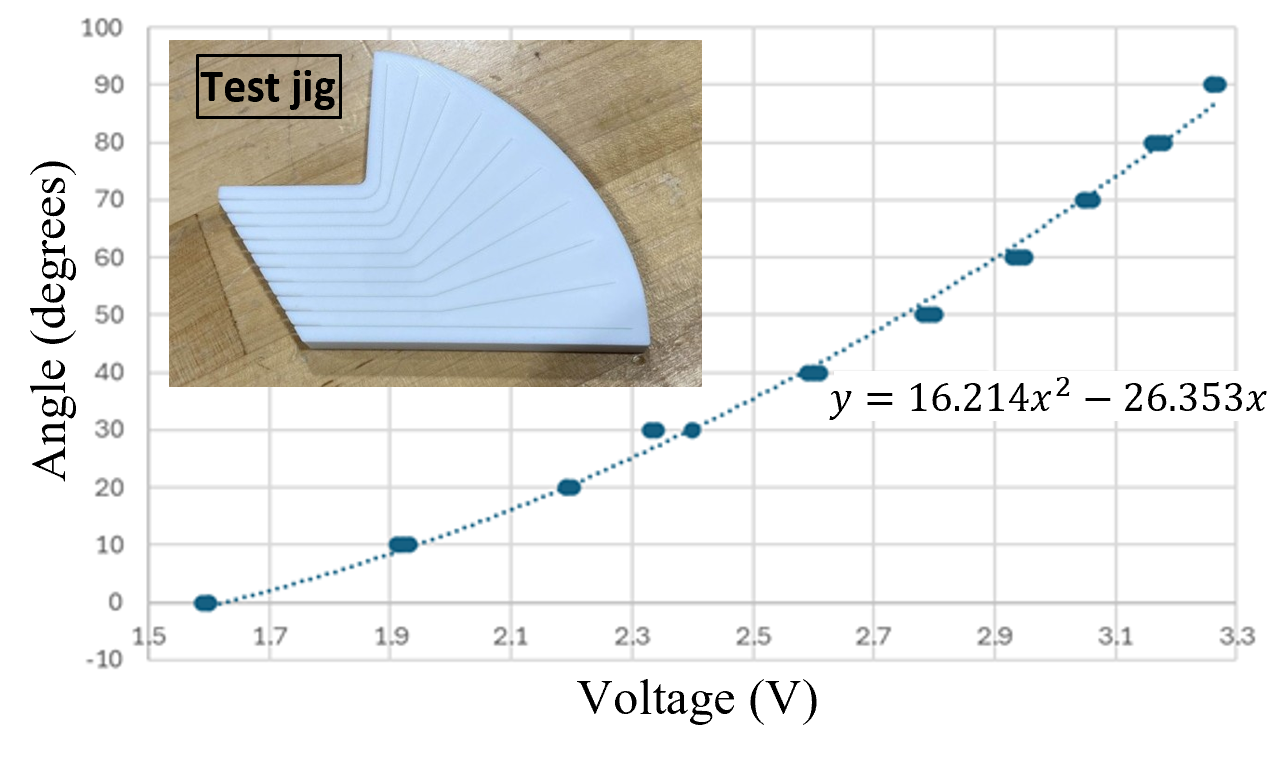}
\caption{Data collected from testing the flex sensor and the resulting fitted function to determine the angle of the flex sensor. \emph{(Top left)} Test jig used for data collection.}
\label{fig.angleTF}
\end{figure}

To calibrate the flex sensor, a test jig was 3D printed (Fig.~\ref{fig.angleTF} \emph{top left}) to measure the resistance of the flex sensor over a bending range from 0-90 degrees in increments of 10 degrees.
The minimum and maximum resistance of the flex sensor were 20 and 60$\text{ k}\Omega$, respectively.
For data acquisition, the flex sensor was connected to a voltage divider circuit.
An Arduino Mega was used for analog-to-digital conversion.
The flex sensor was placed in series with a 47 kOhm resistor to produce a wide range for the voltage output.
The results of testing the flex sensor with the voltage divider are shown in Fig.~\ref{fig.angleTF}.
A second-order polynomial was fit to create a function that converts the voltage readings into angle measurements.
The test jig was used in subsequent data collection to test the accuracy of the calibration, which demonstrated accuracy within +-5 degrees.
The system was more accurate with smaller angles between 0 and 50 degrees and became less accurate with sharper angles.

\subsection{Finite Element Analysis}

\begin{figure}[thpb]
\centering
\includegraphics[width=.9\linewidth]{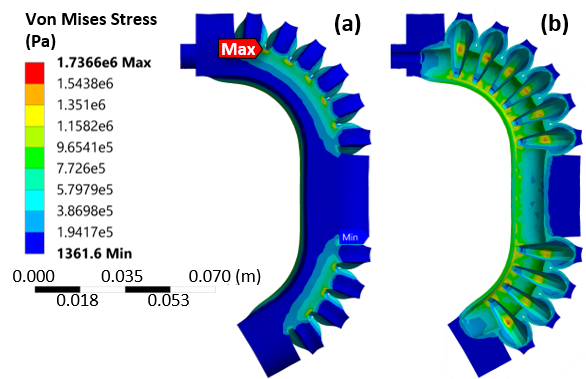}
\caption{Von Mises stress plots from the FEM simulations of the actuator with free bending. 150 kPa of pressure was applied to the internal surfaces of the actuator. (a) Full view. (b) Section view.}
\label{fig.actuatorFEM}
\end{figure}

To analyze the bending behavior of the actuator, FEM simulations were performed using Ansys Workbench.
A static structural analysis was performed with a static pressure load applied as a ramp function from 0 to 150 kPa to the internal faces of the actuator.
The material used to fabricate the actuators, Sorta Clear 40, was modeled using stress-strain data from~\cite{Marechal2021-SoRo}. 
A third-order Yeoh model is used to characterize the hyperelastic material $(C_1=1.00x10^{-1}, C_2=2.11x10^{-1}, C_3=1.66x10^{-3}$).
To account for the expansion of the air chambers, frictionless contacts were added between the external faces of the air chambers.
The results are shown in Fig.~\ref{fig.actuatorFEM}.
The figure shows the von Mises stress plot and the true deformation of the actuator.
The maximum stress occurs on the external faces of the actuator
at the base of the external cuts between the air chambers.
When an internal pressure of 150 kPa is applied, the maximum stress the actuator will experience is $\sim$1.74 MPa, which is well below the 5.52 MPa tensile strength of Sort Clear 40, suggesting safe operation at this level of pressurized deformation.
A section view of the actuator in Fig.~\ref{fig.actuatorFEM}(b) shows the stresses on the internal walls of the silicon body.
This simulation shows that the fPN actuator design is a viable choice because it allows for targeted bending and can withstand large internal pressures to bend the finger. 

\begin{figure*}[thpb]
\centering
\includegraphics[width=.9\linewidth]{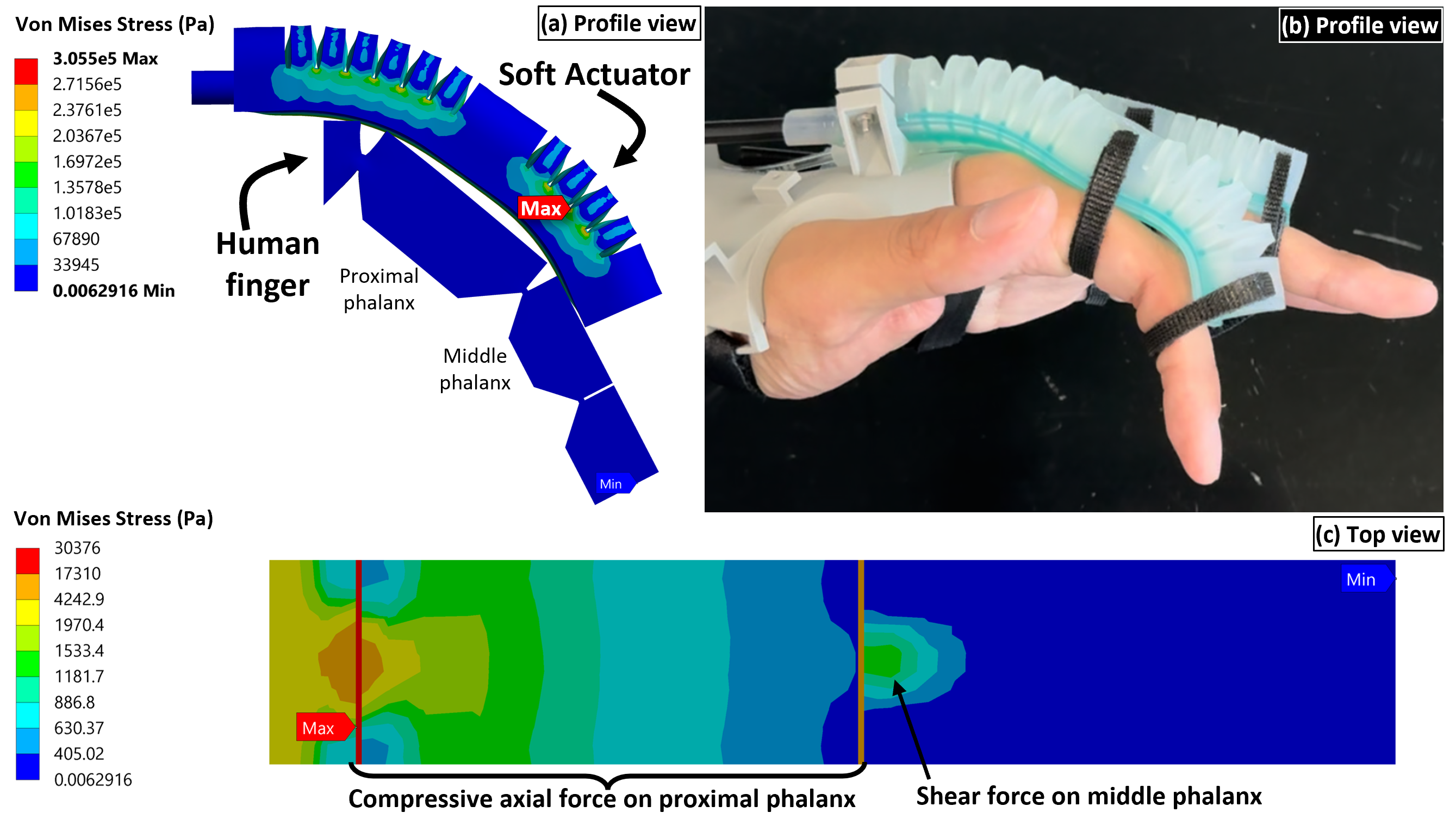}
\caption{Comparison of simulated and real-world experiments of the physical human-robot interaction (pHRI) between a soft actuator and the human index finger. (a) Profile view in the Ansys simulation. The simulation models one soft actuator from the glove being actuated and, in turn, bending the biomechanical model of Participant A's index finger. (b) Physical testing of the actuator with Participant A's finger, from a similar profile view for comparison.
(c) The simulation from a top view, hiding the soft actuator to portray the stresses on the biomechanical finger model, demonstrating non-ideal compressive axial forces on the proximal phalanx and close-to-ideal shear forces on the middle phalanx.}
\label{fig.actuatorTesting}
\end{figure*}

Another static structural analysis was performed to study the physical human-robot interaction (pHRI) when the soft actuator is pressurized to mobilize the finger, as shown in Fig.~\ref{fig.actuatorTesting}a.
A simplified biomechanical finger model was created based on Participant A's dimensions.
The actuator was positioned in the same place that it would be when worn by the participant.
The proximal end of the actuator is fixed.
The simplified biomechanical finger model is a single part with living hinges representing the joints of the finger.
The living hinges are thin pieces of material connecting the phalanges, so when a force is applied to the finger, bending will be focused on the thin connections.
Fig.~\ref{fig.actuatorTesting}a and Fig.~\ref{fig.actuatorTesting}b show side-by-side comparisons of the simulated and real-world experiments from the same profile view.

The material model for the actuator is the same as in the previous free-bending simulation.
The biomechanical finger was modeled with Ecoflex 00-50 material, which has a Shore hardness similar to human flesh.
The Ecoflex was modeled with a third-order Yeoh, with coefficients obtained from~\cite{Kulkarni2015-Thesis}.
$(C_1=1.90x10^{-2}, C_2=9x10^{-4}, C_3=-4.75x10^{-6}$.)
This material allowed for functional joint bending deformation at the living hinges and nominal semi-rigid behavior in the thick solid phalangeal regions.
Frictionless contacts were added between the bottom face of the actuator and the top faces of the finger.
An internal pressure ramp of 0 to 40 kPa (5.8 psi) was applied to the actuator, and the results at the maximum pressure are presented in Figs.~\ref{fig.actuatorTesting}(a,c).
Fig.~\ref{fig.actuatorTesting}a shows a profile view of the von Mises stress plot of the actuator and the simplified finger, with true deformation depicted.
This plot demonstrates how the simplified finger reacts with bending at the joints to the forces of the actuator pressing against it.
Fig.~\ref{fig.actuatorTesting}c shows a top view of the von Mises stress plot of just the human finger model, with the soft actuator hidden, focusing on this lower stress distribution range. 
Fig.~\ref{fig.actuatorTesting}a shows visually that the actuator is applying non-ideal compressive axial forces to the proximal phalanx, and this is depicted in Fig.~\ref{fig.actuatorTesting}c with the higher stress distribution on the proximal phalanx. 
Shear forces on the proximal and middle phalanges are desired in order to cause rotation about the MCP and PIP joints, respectively.  
The inefficiencies observed here serve as a benchmark for future designs.
These simulations are validated through physical testing of the exoglove with the participant's finger (Fig.~\ref{fig.actuatorTesting}a), which shows the index finger actuator pressurized to 20 psi.
Despite differences in the pressure values (20 psi in real-world experiments vs 5.8 psi in simulation), it can be observed that the overall pHRI behavior is comparable.
The pressure value differences can be attributed to the simplification of the human finger model.

\subsection{Personalization Experiments}

\begin{figure}[thpb]
\centering
\includegraphics[width=1\linewidth]{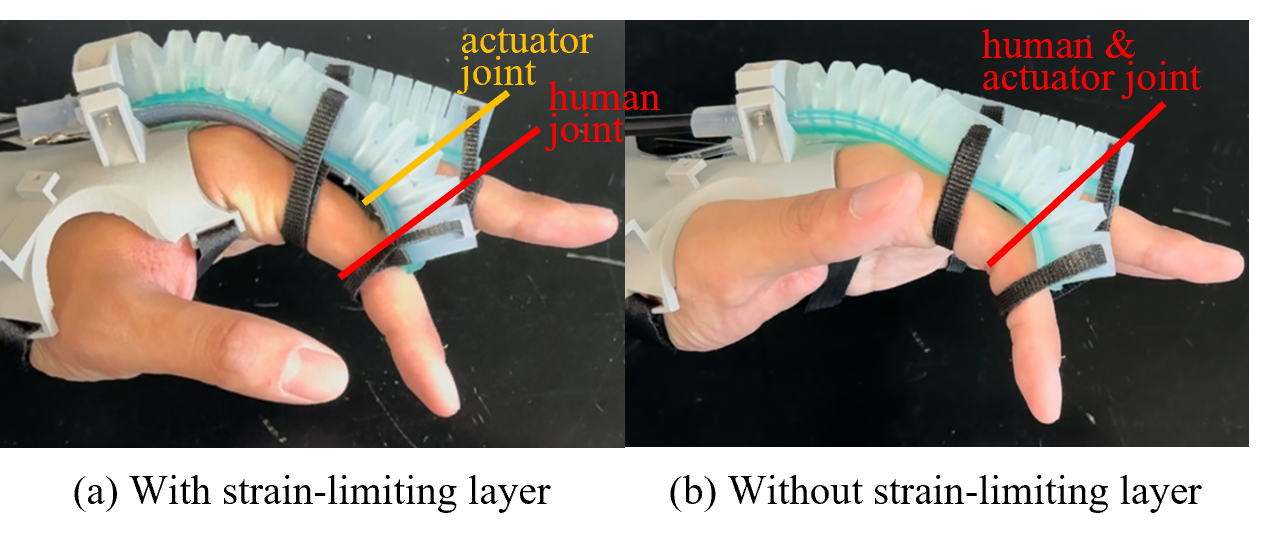}
\caption{Physical testing results of two actuators with different base layers. (a) Actuator with the fabric strain-limiting layer embedded into the base layer. (b) Actuator with nothing embedded in the base layer, allowing for some longitudinal stretching and hence alignment between the actuator and the human PIP joint.}
\label{fig.actuatorBaseTesting}
\end{figure}

While testing the actuators with the participant's hand, we noticed that they would become misaligned with the fingers as they were inflated.
As the finger bends, actuator joints shift away from the finger joints.
This was due to the fabric strain-limiting layer, which prevented the actuator from stretching axially and maintaining proper alignment with the finger joints.
To confirm this, experiments were conducted to test the difference between actuators with and without a strain-limiting layer.
The actuators were fabricated using the same process; for the actuator without the strain-limiting layer, it was made without the fabric and flex sensor.
The results are shown in Fig.~\ref{fig.actuatorBaseTesting}.
In Fig.~\ref{fig.actuatorBaseTesting}a, the test with fabric shows how the PIP joint of the finger (red line) slips past the actuator joint (orange line) as it bends.
However, when there is no fabric, Fig.~\ref{fig.actuatorBaseTesting}b shows that the actuator joint remains aligned with the PIP joint (common red line).
From these tests, it can be concluded that the strain-limiting layer is not required to achieve bending and hinders the performance of the actuator for this application.

\subsection{Pneumatic Control and Actuation}\label{subsec.control}

A closed-loop PID controller was implemented to regulate the internal pressure of each soft pneumatic actuator. 
The pneumatic control station published in~\cite{Massoud2024-IEEEAccess, Massoud2025-RAL} was used for pressure regulation; however, the original 3/2 on/off valve configuration was replaced with two 2/2 proportional valves to reduce pressure fluctuations and improve control smoothness.
Feedback is provided by a pressure sensor, and the controller modulates the pneumatic input to track a desired pressure reference.
As a safety precaution, if the sensor detects that the pressure exceeds a set threshold, the system will fully exhaust the air.
This value was set to 40 psi based on empirical testing of comfort evaluations. 
As a redundant implicit safety assurance, the diaphragm pump being used to generate air pressure is not capable of generating more than 40 psi of pressure.

\begin{figure}[thpb]
\centering
\subfloat[]{\includegraphics[width=.9\linewidth]{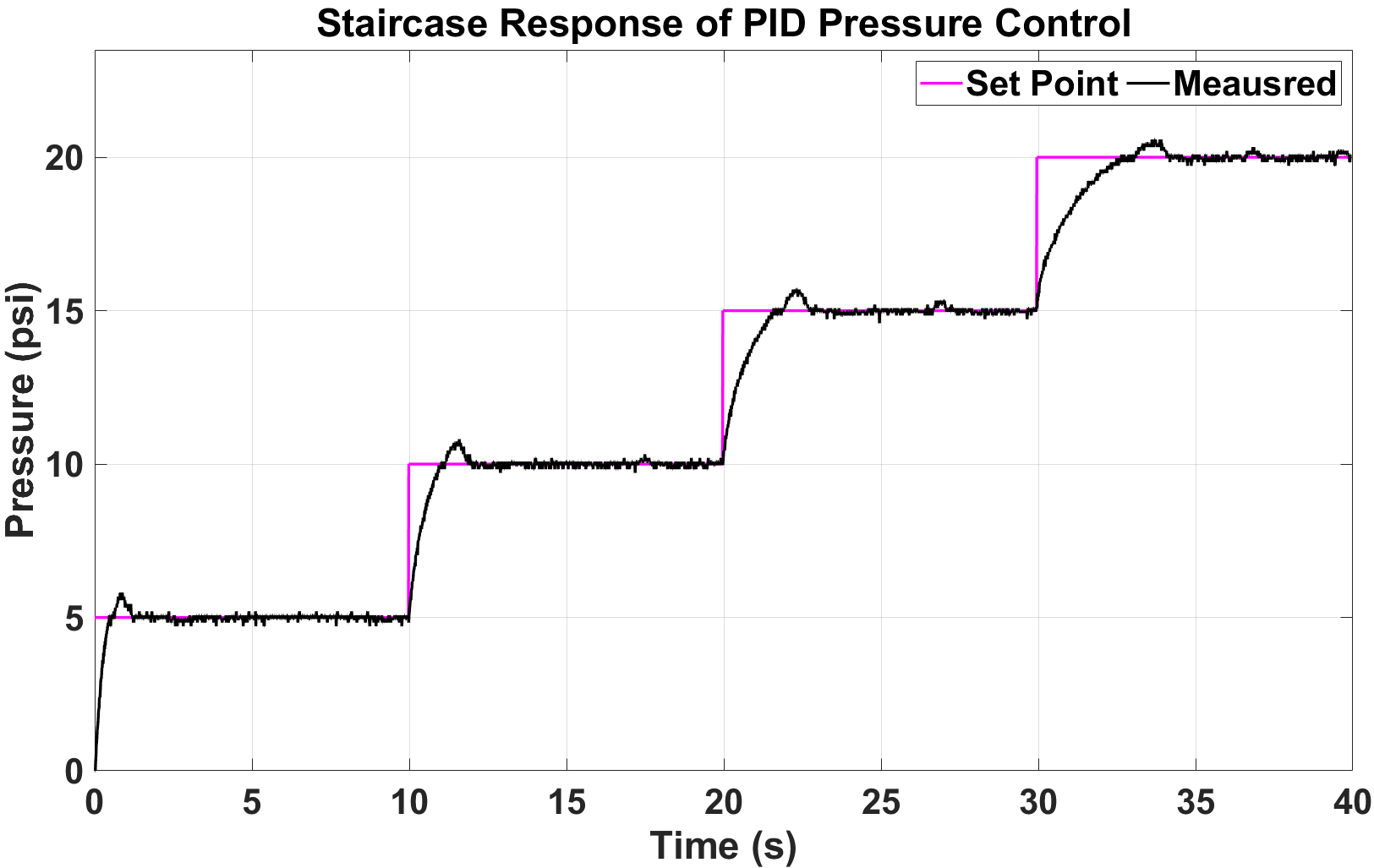}}
\label{fig:figStairCseResponse}
\hfil
\hspace{-0.3cm}
\subfloat[]{\includegraphics[width=.9\linewidth]{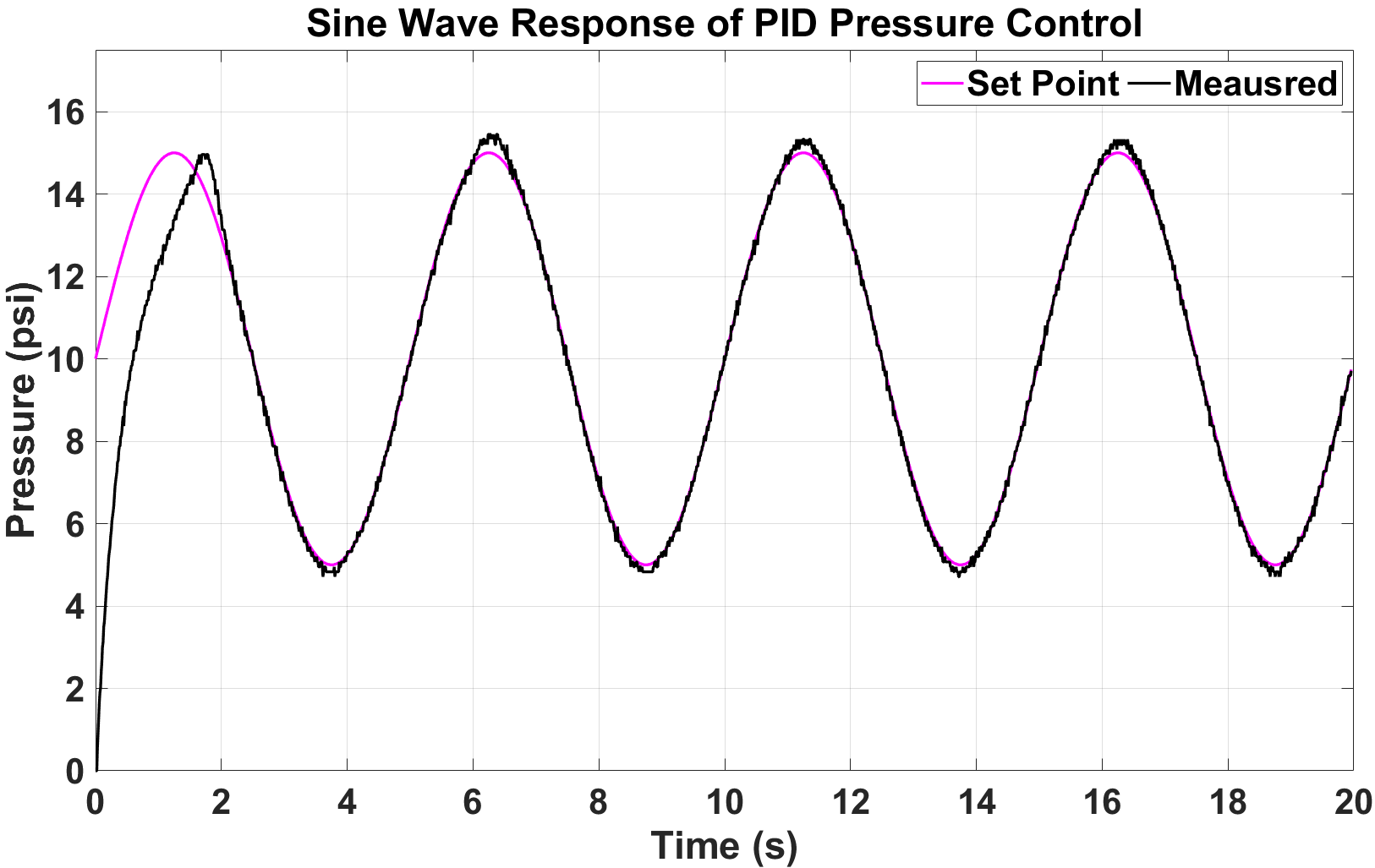}}
\label{fig:figSineWaveResponse}
\subfloat[]{\includegraphics[width=.9\linewidth]{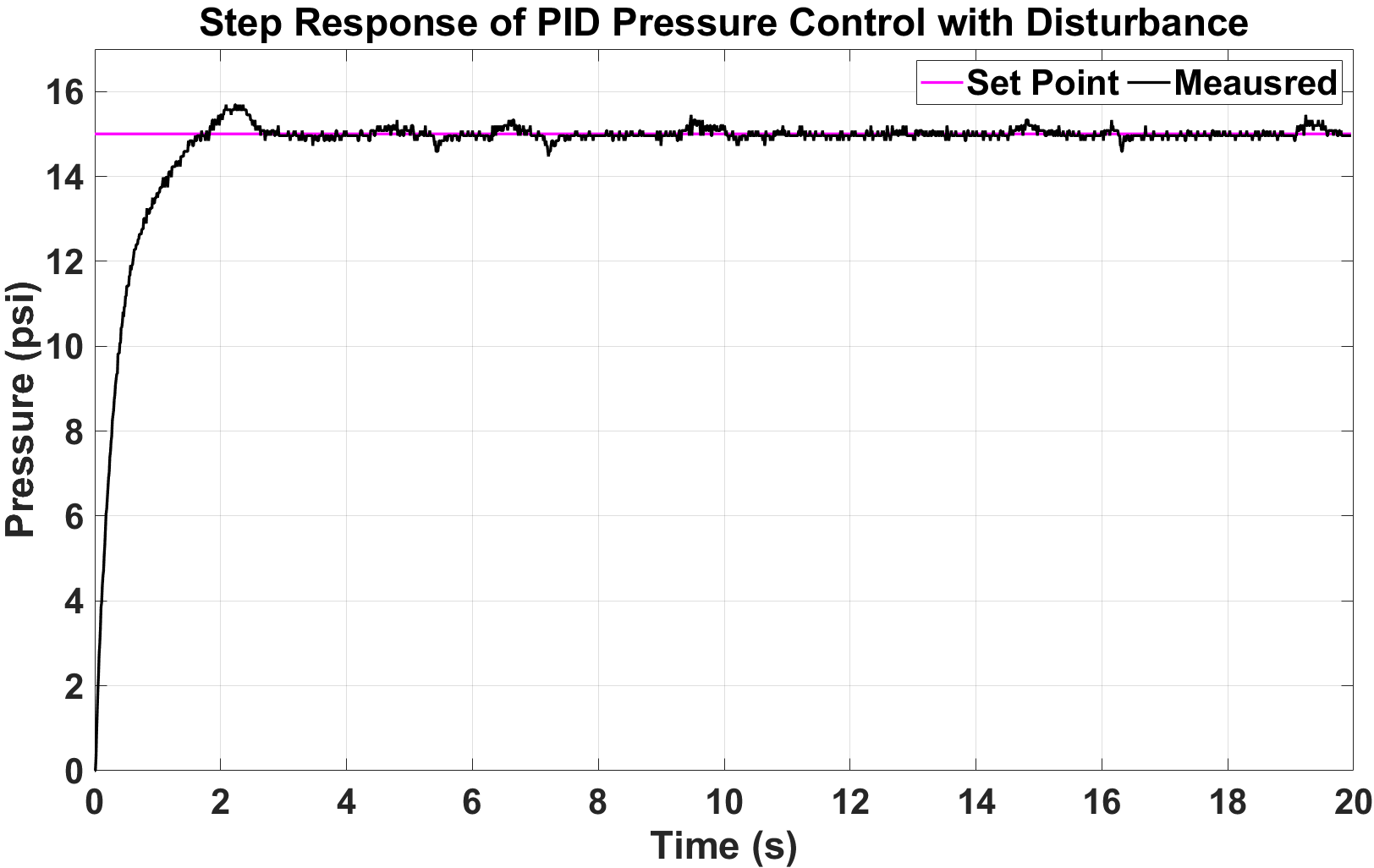}}
\label{fig:figStepWithDisturbance}
\caption{Robotic control experiments with the participant wearing the exoglove. (a) Staircase response with increasing pressure steps. (b) Sine wave response. (c) Step response with disturbance from another person pushing against the participant's finger. (\emph{Refer to supplementary video to view all three experiments.})}
\label{fig:figSineWave}
\end{figure}

The controller gains were tuned empirically to achieve a fast response with minimal overshoot and low steady-state error. The PID gains were set to $K_p = 35$, $K_i = 200$, and $K_d = 0.4$. Fig.~\ref{fig:figSineWave} shows the pressure control performance under different reference inputs. The staircase response in Fig.~\ref{fig:figSineWave}a demonstrates accurate tracking of discrete pressure setpoints, indicating stable regulation across multiple pressure levels. Fig.~\ref{fig:figSineWave}b shows the response to a sinusoidal pressure command, where the measured pressure closely follows the reference with good phase alignment, demonstrating effective dynamic tracking. Robustness to external disturbances was evaluated using a step response, shown in Fig.~\ref{fig:figSineWave}c. 
After the initial rise, the controller rapidly rejects perturbations and maintains the desired pressure.
Perturbations were experimentally introduced by another person pushing against the participant's index finger as the finger was actuated by the glove.
The glove is able to maintain the desired configuration, rejecting the external forces of the second person, while providing compliance to maintain safety and comfort.
The full experiments from Figs.~\ref{fig:figSineWave}a, ~\ref{fig:figSineWave}b, and ~\ref{fig:figSineWave}c can be viewed in the supplementary video.
These experiments highlight the robotic glove's suitability for human–robot interaction scenarios.

The step-response performance is summarized in Table~\ref{tab:step_response}. 
The controller achieved a rise time $t_r$ of 0.937~s and a settling time $t_s$ 
of 1.359~s, indicating a fast transient response. The maximum overshoot 
$M_p$ was 4.60\%, while the steady-state error (SSE) was only 0.0272~psi, 
showing accurate pressure tracking with minimal final error.

\begin{table}[thpb]
\centering
\caption{{Step-response performance.}}
\label{tab:step_response}
\scriptsize
\begin{tabular}{lcccc}
\hline
 & $t_r$ & $t_s$ & $M_p$ & SSE \\
\hline
Value & 0.937 s & 1.359 s & 4.60\% & 0.0272 psi \\
\hline
\end{tabular}
\vspace{-2mm}
\end{table}

\section{CONCLUSIONS}\label{section.conclusions}

This paper presented a pneumatically actuated soft robotic exoglove that was designed to be an anatomically personalized hand rehabilitation device.
Using a 3D scan of a participant’s hand, the pneumatic soft-bending actuators were customized to align with the MCP and PIP joints, and the rigid base was successfully modeled to fit the contours of the dorsal palm.
The FEA simulation of the actuator was used to evaluate its bending behavior, and it was determined that the actuator design is more than robust enough to handle the stresses applied from internal pressurization. 
The simulations of the actuator and the simplified biomechanical model of the participant’s finger provided a framework for analyzing the physical human-robot interaction.
The results indicate that the fPN design can be improved to provide more shear rather than axial forces on the phalangeal regions to more efficiently bend the finger joints.
Physical testing of the exoglove using PID pressure control demonstrated that removing the strain-limiting layer of the actuators improved kinematic alignment when bending, thereby improving on the personalized design.

Pressure control was chosen as a benchmark for future position or force control, since the measurement of air pressure is the most fundamental physical element to sense after air flow, and air pressure sensors are embedded into the operation of the electropneumatic controller.
Since pressure is directly related to contact force, limiting pressure also implicitly limits force, so adding a safety pressure limit as discussed in Section~\ref{subsec.control} also puts an upper bound on the applied force.
The embedded flex sensor will be used for future position control; however, pure position control separate from impedance control could become unsafe.
With pressure control, if the human resists the actuator motion, the pressure will build up and reach its limit; however, with position control, the exoglove would be blind to this resistance.
Although force/torque control is often used for rigid exoskeletons or cable-driven exosuits, it has only been implemented for soft pneumatic exosuits as a binary trigger for position control.  
Future work will examine the use of flexible thin-film force sensors for force control at the contact surface, although commercially available sensors would still be stiffer than the soft actuators themselves, thereby changing the conformal nature of the glove's contact with the human fingers.

Although this article is limited to a pilot study for a single participant, we have outlined steps for a digital manufacturing protocol, enabling future mass customization.
For a new participant, a 3D scan can be taken, which will then allow the conformal customization of the dorsal base, as well as the measurement between the MCP and PIP joints for placement of the corresponding soft actuator joints.
Future work will validate this pipeline with larger participant cohorts. 
Future work will also investigate glove design optimization to minimize unwanted axial compressive forces on the proximal phalanx.

We found that removing the strain-limiting layer improved the exoglove fit, since elongation is necessary for an actuator joint to wrap concentrically around a flexed human joint.
Future work will investigate how to customize the ratio of elongation to rotation, while maintaining structural stability.
In turn, we will investigate how a flex sensor can be embedded into the silicone base while still allowing for some elongation.
Additionally, we will investigate how the simplified biomechanical finger model can be improved to accurately model the forces required to bend the user’s fingers.

This paper presents a promising step towards soft robotic exogloves that can provide high degree of freedom, personalized joint articulation to enable assistance with fine motor skills in the hand.  This technology can enable future assistance for activities of daily living, as well as neuro-rehabilitation of fine motor skills, potentially leading to a new level of neuroplasticity regeneration.



\section*{ACKNOWLEDGMENTS}

\bibliographystyle{unsrt}
\bibliography{references}

\end{document}